\documentclass{article}

\usepackage{times,latexsym}
\usepackage{url}
\usepackage{algorithm}
\usepackage{algpseudocode}
\usepackage{amssymb}
\usepackage{amsmath}
\usepackage{amsthm}
\usepackage{authblk}
\usepackage{booktabs}
\usepackage{graphicx}
\usepackage{xcolor}
\usepackage{hyperref}
\usepackage{tcolorbox}
\usepackage{adjustbox}
\usepackage{multirow}

\begin{document}
\title{LCoT-GV: Graph Attention Networks for Verifying Long Reasoning Chains in Large Language Models} 
\author[1,2]{Bérénice Jaulmes}
\author[1]{Mehwish Alam}

\affil[1]{Télécom Paris, Institut Polytechnique de Paris, France}
\affil[2]{BNP Paribas, IT Group, France}

\date{\today}
\maketitle

\begin{abstract}
Large Reasoning Models produce Long Chains-of-Thought (LCoTs) which involve breaking down the problem into smaller reasoning steps before reaching the conclusion. However, these steps often contain contradictions, unsupported inferences, or irrelevant steps, even when the final answer is correct.
We propose Long Chain-of-Thought Graph Verifier (LCoT-GV), a graph-based framework that represents LCoTs as reasoning graphs. Each node in the graph represents a reasoning step and the edges encode semantic and logical relations. A Graph Attention Network is then trained to predict chain-of-thought correctness from the reasoning graph. We construct a new graph-oriented verification dataset from multiple reasoning benchmarks for question answering in various domains. The results show that our method is competitive with the most similar approaches.
\end{abstract}

\section{Introduction}
\label{sec:introduction}

Long Chains-of-Thoughts (LCoTs), produced by Large Reasoning Models (LRMs)~\cite{chen2026towards}, introduce various challenges due to the complexity and length of the generated reasoning chains. Different steps within the same chain may contradict one another, contain arithmetic inconsistencies, introduce irrelevant information, or rely on unsupported inferences while still leading to a correct final answer~\cite{vacareanu2024general}. Existing approaches for verifying reasoning chains generally analyze reasoning sequentially or locally. Graph-based approaches like \cite{fang2026graph} leverage the structural relations present in the reasoning processes. Most of these verification methods, such as LCoT2Tree~\cite{jiang2025makes}, are also generally dependent on the use of LLMs to build their representations, making the verification process expensive. 

In this work, we propose LCoT Graph Verifier (LCoT-GV), a framework for graph-based verification of reasoning chains generated by LRMs. We represent LCoTs as graphs where nodes correspond to reasoning steps and edges encode semantic and logical dependencies identified using a Natural Language Inference (NLI) model. 
Graph Attention Networks (GATs) are then used for verifying if a reasoning process is correct or incorrect. As compared to other methods, LCoT-GV constructs the reasoning graph entirely locally instead of requiring multiple calls to LLMs, which reduces computational cost.

Although the DeltaBench~\cite{he2025largelanguagemodelsdetect} dataset contains evaluated LCoTs, it contains too few samples to train LCoT-GV. We therefore introduce a new dataset specifically designed for graph-oriented reasoning verification, containing LCoTs, correctness labels, and graph structures derived from semantic relations between reasoning steps.

The experimental results demonstrate that LCoT-GV achieves performance competitive with the most similar graph-based method LCoT2Tree.  Our results further indicate that graph structure alone can provide valuable information for certain downstream tasks, although incorporating semantic information is necessary to achieve further performance improvements. The effectiveness of LCoT-GV also varies across downstream tasks, highlighting the importance of task-specific characteristics. Finally, the model used to generate the LCoT influences performance, although this effect is comparatively smaller than that of the downstream task. The code is available at \footnote{https://github.com/ormarv/LCoT-GV}.

\section{Related Work}
\label{sec:related-work}

Graph-of-Verification~\cite{fang2026graph} represents a CoT as a graph with different node levels. Each node is verified, starting from the root, and verification stops in a branch if an error is detected.
ReasoningFlow~\cite{lee2025reasoningflow} presents an annotation scheme that defines nine different types of steps and three types of relations between them. 

Thinking Reward Model~\cite{zhang2026characterizing} relies on a graph representation of a CoT to evaluate the step- and chain-level quality of the reasoning. However, these methods are mostly focused on CoT instead of LCoT.
LCoT2Tree~\cite{jiang2025makes} represents LCoT as a tree. This tree is built by using an LLM to map each thought segment to an abstract depth index from an extracted task sketch. A thought is inserted as a child node if its index advances past the current node, or attached higher up the tree under a previous parent if its index indicates backtracking or a branch reset. For further details please refer to~\cite{jaulmes2026survey}. 

In contrast, our work proposes a graph-based framework for reasoning verification in which reasoning steps are represented as nodes connected through semantic and logical relations identified using NLI models instead of LLMs. 
\section{LCoT-GV}
\label{sec:methodology}

\begin{figure*}[t]
    \centering
    \includegraphics[width=\textwidth]{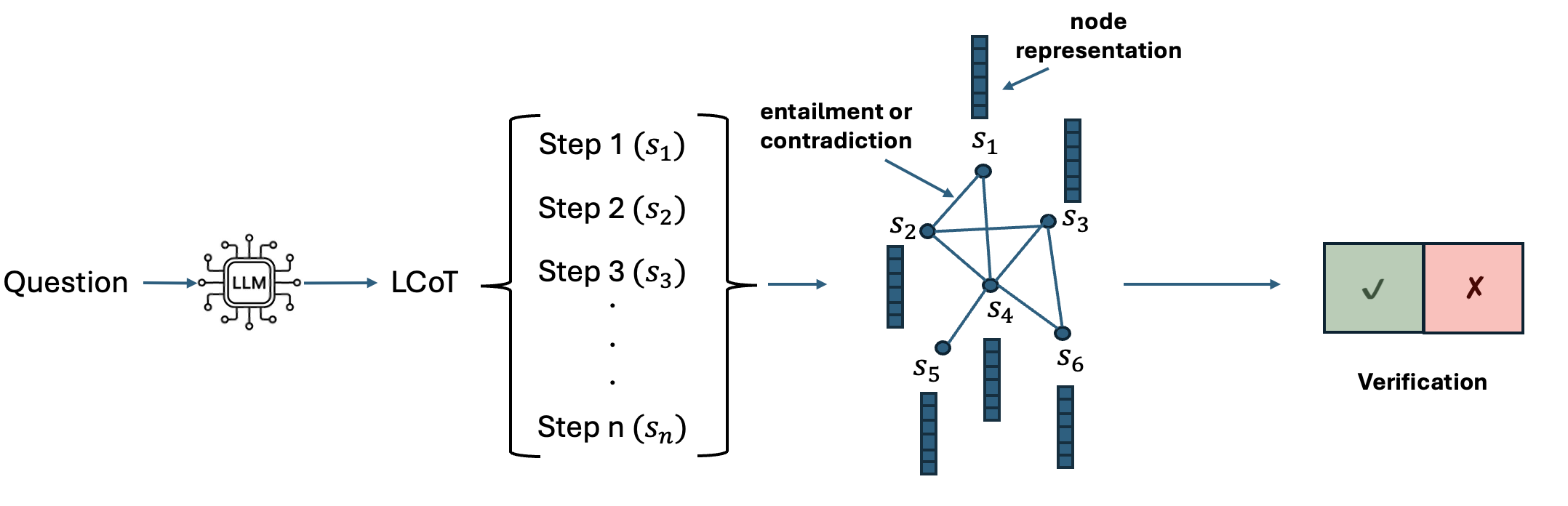}
    \caption{Overview of the proposed LCoT-GV pipeline.}
    \label{fig:pipeline}
\end{figure*}

Figure~\ref{fig:pipeline} shows the overall architecture of Long Chain-of-Thought - Graph Verifier (LCoT-GV), we construct a graph from each LCoT split into steps. We then learn graph embeddings based on these reasoning graphs. These representations are further used for determining if the step is correct.

\paragraph{Constructing Reasoning Graphs.}

Each step in the LCoT is split on important keywords, such as \emph{"So", "Actually", "Let's"}, or \emph{"Wait"}, rather than double newline~\cite{zhang2026characterizing}. This creates steps that are more coherent and meaningful because these keywords indicate logical transitions between steps. Each step is then represented as a node in the reasoning graph.
The edges between the nodes represent entailment or a contradiction among the steps using NLI.

When inserting a new step, we first determine the pool of potential parents (see Algorithm~\ref{alg:getcandidateparents}).
The immediate previous node is examined to identify the branch where it was inserted. 
This branch is designated as the \emph{main branch}. For each node in the main branch, we compute the number of children linked by an \emph{Entailment} relation (line 5). A pool of candidate parent nodes is then constructed by including \textbf{(i)} the last 30 nodes from the main branch (or all nodes if the branch contains fewer than 30), \textbf{(ii)} all earlier nodes on the main branch that have more than one positive child, and all leaf nodes (i.e., nodes without children) in the graph. 
For each candidate parent, the last five nodes from the branch on which it was inserted are collected, and their corresponding text is used as the context for comparison.

The NLI classifier compares this context against the new node being inserted and predicts either ``entailment" (positive edge) or ``contradiction" (negative edge). Positive and negative edges are created when the classifier score exceeds predefined thresholds. Among all candidate parents, the branch associated with the highest entailment score is stored as the \emph{main branch} for the newly inserted node. The three candidate parents with the highest entailment scores, and the two candidate parents with the highest contradiction scores, are linked to the current node with positive and negative edges respectively.

LRMs may occasionally produce the same reasoning step or sequence of steps hundreds or even thousands of times. As a result, the generated LCoT can contain thousands of nearly identical steps, making graph construction computationally expensive. To mitigate this issue, we detect repeated subsequences and reuse the graph structure of their original occurrence instead of reconstructing it from scratch.

\begin{algorithm*}
    \caption{Identify Candidate Parents}
    \label{alg:getcandidateparents}
    \begin{algorithmic}[1]
        \Require Graph $G$, main\_branch, last inserted node $i$
        \State $leaves \gets \texttt{getLeaves}(G)$ 
        \State $candidates \gets \emptyset$
        \State $candidates \gets candidates \cup leaves$
        \For{$idx$, $n \in \texttt{enumerate}(\textit{main\_branch})$}
            \State $positiveChildren \gets $ {$j$ | $(s_{n}, s_{j}) \in E \land label((s_{n}, s_{j})) = \text{`Entailment'}$}
            \If{($i\geq \texttt{lengthOf}(\textit{main\_branch})-30) \lor (|positiveChildren| \geq 2)$}
                \State $candidates \gets candidates \cup {n}$
            \EndIf
        \EndFor
        \State Sort $candidates$ in descending order
        \State \Return $candidates$
    \end{algorithmic}
\end{algorithm*}

\paragraph{Learning Representations from Reasoning Graphs.}
For the node features, we chose to use the embeddings of the step corresponding to each node within the reasoning graph, to learn from their rich semantic information.
These embeddings are created using a Sentence Transformer model.
 
A GAT is trained to predict the correctness of the final answer produced by the LCoT reasoning process based on its corresponding reasoning graph. We optimize the model using a negative log-likelihood loss. The GAT takes three inputs: node features, edge features, and an adjacency matrix. Each edge is represented by a one-hot encoding of the relation between its two connected nodes.
The precise implementation of the GAT model is described in Section~\ref{sec:experimental-setup}.

The Graph Attention layers learn a representation of the reasoning graph, which is subsequently passed to a linear classification layer. This final layer performs graph-level classification, predicting the correctness of the final answer.

\section{Experimentation}
\label{sec:experiments}

\paragraph{Datasets.}
\label{sec:datasets}
We constructed graph-based representations from a dataset of 8,000 evaluated LCoTs.

Following~\cite{jiang2025makes}, we compiled an 8,000-sample dataset by evenly sampling 2,000 LCoTs from four benchmarks:MMLUpro~\cite{wang2024mmlu}, MATH~\cite{hendrycks2021measuring}, LiveCodeBench-v5~\cite{jain2025livecodebench}, and GPQA~\cite{rein2024gpqa}\footnote{MMLUpro is licensed under Apache 2.0; MATH, LiveCodeBench-v5, and GPQA are licensed under MIT.}
The dataset is fully balanced across three generating LRMs as well as correct and incorrect final answers.
We evaluated outcomes task-specifically:regex matching for MCQs (MMLU-Pro, GPQA),a specialized LaTeX and mathematical expressions parser \footnote{https://github.com/hendrycks/math} for MATH, and the official execution library\footnote{https://github.com/LiveCodeBench/LiveCodeBench/} for LCB.

We split this dataset between the train and test sets (80\% and 20\%, respectively). We use 10\% of the train set for validation.
Our final graph dataset is obtained by generating a graph representation for each LCoT in this dataset.

\paragraph{Experimental Setup.}
\label{sec:experimental-setup}
 
For experimentation, we chose three LRMs: two distilled versions of DeepSeek-R1~\cite{deepseekai2025deepseekr1incentivizingreasoningcapability}: DeepSeek-R1-Distill-Llama-70B and DeepSeek-R1-Distill-Qwen-32B, and QwQ-32B~\cite{qwq32b}. We specifically selected open-source LRMs that could be deployed on one or two H100 GPUs while still providing strong reasoning capabilities.

Our pipeline splits steps using the eight most frequent keywords. We use a long-context DeBERTa~\cite{laurer2024less} for NLI (using 0.7 as threshold for both entailment and contradiction) and a contrastively fine-tuned MiniLM~\cite{wang2020minilm} for sentence embeddings.

Our GAT model consists of two GATv2 layers (hidden dimension 64) and a two-layer MLP classification head with ReLU activation. We train for 100 epochs using a batch size of 32 and a learning rate of 1e-3.

Generating 8,000 LCoTs required approximately 16 hours on two H100 GPUs. Graph construction took up to 3 minutes per sample on one V100, and GAT training took under 30 minutes on a V100.

\begin{table*}[t]
    \centering
    \small
    \setlength{\tabcolsep}{6pt}
    \resizebox{\textwidth}{!}{
    \begin{tabular}{llccccc}
    \toprule
        & & MATH & GPQA & LiveCodeBench & MMLU-Pro & Average over datasets \\
    \midrule
        \multirow{3}{*}{DeepSeek-R1-Distill-Llama-Qwen-32B} 
        & Length    & \underline{74.13} & 67.08 & 81.59 & 59.95 & 66.27 \\
        & LCoT2Tree & \textbf{80.81} & \underline{70.37} & \underline{82.21} & \textbf{72.41} & \textbf{75.39} \\
        & Ours      & 69.92 & \textbf{76.82} & \textbf{88.79} & \underline{65.42} & \underline{75.24} \\
    \midrule
        \multirow{3}{*}{DeepSeek-R1-Distill-Llama-70B} 
        & Length    & -     & -     & -     & -     & -     \\
        & LCoT2Tree & -     & -     & -     & -     & -     \\
        & Ours      & 76.54 & 78.75 & 85.17 & 71.23 & 77.92 \\
    \midrule
        \multirow{3}{*}{QwQ-32B} 
        & Length    & \underline{75.82} & 62.09 & 78.30 & 58.00 & 66.97 \\
        & LCoT2Tree & \textbf{77.63} & \underline{68.55} & \underline{80.05} & \textbf{72.59} & \underline{73.96} \\
        & Ours      & 70.64 & \textbf{79.85} & \textbf{88.77} & \underline{72.40} & \textbf{77.92} \\
    \midrule
        Average over models & Ours & 72.37 & 78.47 & 87.58 & 69.68 & 77.03 \\
    \bottomrule
    \end{tabular}
    }
    \caption{Accuracy across LRMs and benchmarks, averaged over five runs.}
    \label{tab:main_results}
\end{table*}

\begin{table*}[h]
    \centering
    \small
    \setlength{\tabcolsep}{6pt}
    \resizebox{\textwidth}{!}{
    \begin{tabular}{llccccc}
    \toprule
        & & MATH & GPQA & LiveCodeBench & MMLU-Pro & Average over datasets \\
    \midrule
        \multirow{3}{*}{DeepSeek-R1-Distill-Llama-Qwen-32B} 
        & Default model    & \textbf{69.92} & 76.82 & \underline{88.79} & \textbf{65.42} & \textbf{75.24} \\
        & Meta & 47.77 & 56.69 & 66.51 & 40.71 & 52.92 \\
        & Mixed & 64.60 & \textbf{78.62} & 88.09 & 61.86 & 73.29 \\
        & Mixed, P & 66.10 & \underline{78.04} & \textbf{89.58} & \underline{64.50} & \underline{74.55} \\
        &Embeddings, P & \underline{67.81} & 77.77 & 85.63 & 62.36 & 73.39 \\
    \midrule
        \multirow{3}{*}{DeepSeek-R1-Distill-Llama-70B} 
        & Default model & \textbf{76.54} & \textbf{78.75} & 85.17 & \textbf{71.23} & \textbf{77.92} \\
        & Meta & 45.32 & 48.38 & 69.90 & 49.67 & 53.32 \\
        & Mixed & 65.57 & \underline{73.52} & \textbf{88.29} & 68.69 & 75.24 \\
        & Mixed, P & 65.52 & 74.11 & 86.11 & 69.60 & 73.84 \\
        &Embeddings, P & \underline{70.70} & 73.46 & \underline{86.83} & \underline{71.02} & \underline{75.50} \\
    \midrule
        \multirow{3}{*}{QwQ-32B} 
        & Default model    & \underline{70.64} & \textbf{79.85} & \underline{88.77} & 72.40 & \textbf{77.92} \\
        & Meta & 52.27 & 51.21 & 65.94 & 50.25 & 54.92 \\
        & Mixed & 66.52 & \underline{77.41} & \textbf{88.91} & 71.12 & 75.99 \\
        & Mixed, P & 67.51 & 77.07 & 90.53 & \textbf{73.63} & 77.18 \\
        &Embeddings, P & \textbf{72.90} & 76.56 & 88.15 & \underline{73.42} & \underline{77.76} \\
    \bottomrule
    \end{tabular}
    }
    \caption{Accuracy across LRMs and benchmarks for different model variants, averaged over five runs. Meta: metadata used as features; Mixed: embeddings and metadata jointly used as features; P: positive edges only.}
    \label{tab:variation_results}
\end{table*}

\paragraph{Results.}
\label{sec:results}

We compare our model against LCoT2Tree, the approach most close to our work, and a length-based classifier. Table~\ref{tab:main_results} shows that LCoT-GV achieves an average score of 76.58 across the two LRMs outperforming LCoT2Tree (74.68). There are strong gains on LCB (+6.58 to +8.72) and GPQA (+6.45 to +11.30), as code and scientific concepts closely resemble natural language, yielding better representations for NLI. Conversely, performance drops on MATH (-10.89 to -6.99) and MMLU (-6.99 to -0.19), because language models struggle with mathematical inference \cite{de2025math}.

\paragraph{Model Variations.}
\label{sec:variations}
We tested the following variations of LCoT-GV. Table~\ref{tab:variation_results} shows the results for each of the variations described below:

\noindent \textbf{1. Meta-data Embeddings only}: Replaces semantic embeddings with structural metadata (number of children/parents, chain index, nodes on the same level, and LCoT proportion before/after step generation). As a result, the performance falls to random guessing except on LCB, proving semantic information drives most of the model's success.

\noindent \textbf{2. Both feature types}: Processes both feature sets through added linear layers before concatenating them into the first graph attention layer. The overall performance decreases, however, the performance on LCB improves (+1.16).

\noindent \textbf{3. No negative edges}: Removes the negative edges from the base model. The performance drops (from -0.16 to -2.42), confirming their small positive effect.

\noindent \textbf{4. Both feature types, no negative edges}: Compared to the standard combined-feature model, this improves scores for DeepSeek-R1-Distill-Qwen-32B (+1.26) and QwQ-32B (+1.19), but performance drops on Llama (-1.30).

\section{Conclusion}
We introduced ``LCoT-GV", a graph-based framework for verifying LCoT reasoning. By leveraging a local NLI model in place of computationally expensive LLM calls, LCoT-GV reduces computational overhead while enabling improved scalability. Empirical evaluation shows that LCoT-GV outperforms the most closely related existing method on average. Although the verification of mathematical language remains challenging, LCoT-GV achieves particularly strong performance on coding tasks, highlighting its effectiveness in domains with structured reasoning.

\newpage

\section*{Limitations}
\label{sec:limitations}
To facilitate direct comparison with the most closely related existing method \cite{jiang2025makes}, we adopted its data collection procedure for obtaining LCoTs. While this choice ensures methodological comparability, it also constrains the diversity of downstream tasks represented in our evaluation. In future work, we plan to broaden the scope of our evaluation by focusing on non-mathematical reasoning tasks, where our approach may be better suited to LCoT verification.

\section*{Acknowledgements}
The authors thank the Grand Équipement National de Calcul Intensif (GENCI) for providing the necessary computing resources for this project.
\bibliographystyle{plain}
\bibliography{biblio}

\end{document}